\documentclass[a4paper,conference]{IEEEtran}

\usepackage{amsmath,graphicx}
\usepackage{amssymb}
\usepackage{subcaption, xcolor, multirow}
\usepackage{cite, hyperref}

\def\x{{\mathbf x}}
\def\y{{\mathbf y}}
\def\n{{\mathbf n}}

\def\c{{\mathbf c}}

\def\hatx{\mathbf{\hat{x}}}
\def\hatc{\mathbf{\hat{c}}}

\definecolor{mypink}{RGB}{236, 0, 141}
\begin{document}

\title{Deep Universal Blind Image Denoising}

\author{\IEEEauthorblockN{Jae Woong Soh}
\IEEEauthorblockA{Department of ECE, INMC \\
	Seoul National University\\
	Seoul, Korea\\
Email: soh90815@ispl.snu.ac.kr}
\and
\IEEEauthorblockN{Nam Ik Cho}
\IEEEauthorblockA{Department of ECE, INMC \\
	Seoul National University\\
	Seoul, Korea\\
	Email: nicho@snu.ac.kr}}

\maketitle

\begin{abstract}
Image denoising is an essential part of many image processing and computer vision tasks due to inevitable noise corruption during image acquisition. Traditionally, many researchers have investigated image priors for the denoising, within the Bayesian perspective based on image properties and statistics. Recently, deep convolutional neural networks (CNNs) have shown great success in image denoising by incorporating large-scale synthetic datasets. However, they both have pros and cons. While the deep CNNs are powerful for removing the noise with known statistics, they tend to lack flexibility and practicality for the blind and real-world noise. Moreover, they cannot easily employ explicit priors. On the other hand, traditional non-learning methods can involve explicit image priors, but they require considerable computation time and cannot exploit large-scale external datasets. In this paper, we present a CNN-based method that leverages the advantages of both methods based on the Bayesian perspective. Concretely, we divide the blind image denoising problem into sub-problems and conquer each inference problem separately. As the CNN is a powerful tool for inference, our method is rooted in CNNs and propose a novel design of network for efficient inference. With our proposed method, we can successfully remove blind and real-world noise, with a moderate number of parameters of universal CNN.
\end{abstract}


%
\IEEEpeerreviewmaketitle

\section{Introduction}
\label{sec:intro}

Image denoising aims to recover the latent clean image from an observed noisy image. It has been a longstanding fundamental problem because noise intervention is inevitable during the image acquisition process, which degrades the visual quality of the acquired image and lowers the performance of computer vision tasks. The overall noise is the accumulation of multiple different noise sources, such as capturing sensors, in-camera pipeline, and data transmission media. Such a noise generation process is too complicated to address, therefore with the belief of the central limit theorem, the noise is usually assumed as additive white Gaussian noise (AWGN).

Specifically, an observed noisy image $\mathbf{y}$ has been assumed to be the sum of the latent clean image $\mathbf{x}$ and an AWGN $\mathbf{n}$ as
\begin{equation}
\y=\x+\n.
\end{equation}
Traditionally, image denoising has been addressed as a statistical inference problem, which focused on constructing a maximum a posteriori (MAP) inference model. In the Bayesian perspective, the MAP inference can be divided into the data fidelity term (log-likelihood) and regularization term (log-prior) as
\begin{align}
\hatx & = \arg \max_{\x} \log p(\x|\y),\\
&=\arg \max_{\x} \log p(\y|\x) + \log p(\x).
\end{align}
To obtain a plausible model in this form and its solution, many classical researches focused on the design of image prior terms based on experience and knowledge of the data. These approaches include total variation regularization \cite{tv1}, sparsity \cite{K-SVD}, low-rank constraints \cite{WNNM, TWSC}, non-local self-similarity \cite{non-local1, non-local2, non-local3, BM3D}, and so on. These methods \emph{explicitly} defined the data fidelity term with the assumption of AWGN and appropriate image prior term. They are shown to be superior in terms of interpretability and flexibility, but they are limited when the noise deviates from spatially uniform i.i.d. Gaussian noise. Also, their optimization implementation mostly depends on iterative solvers, which takes a long computation time.

Recently, the upsurge of powerful deep learning in computer vision introduced deep learning-based denoisers, which \emph{implicitly} learns the MAP estimate based on the supervised learning framework \cite{DnCNN, FFDNet, RED, Non-localCNN, UNLNet, ATDNet, NLRN, RNAN, VDN}. These methods surpass the conventional non-learning methods by large margins, but they lack in flexibility, especially for blind denoising. For example, if a network is trained to denoise a specific level of noise variance, the network would not work well for an unseen noise-level. Hence, a universal network is needed, which is trained with the whole expected range of noise-level for blind denoising unless the noise-level is known. In this case, it is still inferior to the specific noise-level denoiser, and discriminative learning methods suffer from the averaging problem, which means that the network targets better to the mid-level noise \cite{DnCNN, FFDNet}.
Moreover, knowledge from human experience cannot be easily injected into the deep learning framework. Although there are some works to merge non-local self-similarity, one of the strong priors, with deep networks \cite{Non-localCNN, NLRN, RNAN}, introducing other priors is not easy.

In this paper, we propose a convolutional neural network (CNN)-based universal blind denoiser, dubbed Deep Universal Blind Denoiser (DUBD), which leverages the advantages of MAP inference and the power of deep-learning. The design of DUBD is motivated by the advantages and disadvantages of the above-stated approaches, how to insert human knowledge priors to the deep-learning framework without harming the power of the network. In particular, for practicality, the proposed DUBD can handle a wide range of noise-level and spatially varying noises without knowing the noise-level. Also, our method can address spectrally variant noise, as evidenced in \cite{TWSC}. Moreover, our DUBD outperforms other methods, including blind and non-blind denoisers, while requiring a comparable number of parameters. Our contributions can be summarized as:

\begin{itemize}
\item We propose a CNN-based universal blind denoiser that can handle a wide range of noise-level including spatially and spectrally varying noise.
\item Our DUBD can explicitly incorporate prior knowledge, which further lifts the performance of the network.
\item Our DUBD outperforms other denoisers with a comparable number of parameters, which eventually brings better practicality.
\item Our DUBD can be applied to real-world noisy images and also shows outstanding performance compared to the other methods.
\end{itemize}

\section{Probabilistic View}
\label{sec:motivation}

For blind image denoising, a na\"{i}ve approach is to train a universal CNN with the whole expected range of noise-level. However, it is proven not to be a good choice, lacking performance and flexibility \cite{DnCNN, FFDNet}. Hence, we approach the blind image denoising problem with the divide-and-conquer scheme. Precisely, we reformulate the log-posterior by introducing a new random variable $\c$ that contains prior based on human knowledge as

\begin{align}
\label{eq:4}
\log p(\x|\y) &= \log \int_{\c} p(\x|\y,\c) p(\c|\y) d\c, \\
\label{eq:5}
&\approx \log p(\x|\y,\hatc) p(\hatc|\y),\\
\label{eq:6}
&= \log p(\x|\y, \mathbf{\hat{c}}) + \log p(\mathbf{\hat{c}|y}).
\end{align}
Note that Eq.~\ref{eq:4} is a marginal expression by introducing a random variable $\c$, but the integration is mostly not tractable. Hence, for approximating the integration, we use the point estimate for $\c$, which is $\hatc=\arg \max_{\c} p(\c|\y)$. When $p(\c|\y)$ has a unimodal distribution with a sharp peak, the approximation is quite fair. Then, we can solve the MAP estimate with the given point estimate $\hatc$. Formally, the MAP inference problem can be reformulated into solving two sub-problems as 
\begin{align}
\label{eq:7}
\mathbf{\hat{c}} &= \arg \max_{\c} \log p(\c|\y),\\
\label{eq:8}
\hatx &= \arg \max_{\x} \log p(\x|\y, \mathbf{\hat{c}}).
\end{align}
As neural networks are experts at inference, we employ CNNs for our two inference problems, described as
\begin{align}
\mathbf{\hat{c}} &\approx g_\theta(\y),\\
\hatx &\approx f_\phi(\y;\mathbf{\hat{c}}),
\end{align}
where $g_\theta(\cdot)$ and $f_\phi(\cdot)$ are CNNs with parameters $\theta$ and $\phi$, respectively.
By dividing the original problem into sub-problems, we decrease the modalities of multi-modal distribution of the log-posterior. Thus the network conquers separate conditionals and eventually facilitates solving the whole blind denoising inference. Concretely, for solving the second inference problem, we design a tunable CNN according to $\c$. Importantly, we must introduce an appropriate $\c$ based on our prior knowledge.

\section{Conditional Estimation Network}
\label{sec:CENet}

\begin{figure}[t]
	\begin{center}
		\centering
		\includegraphics[width=1.0\linewidth]{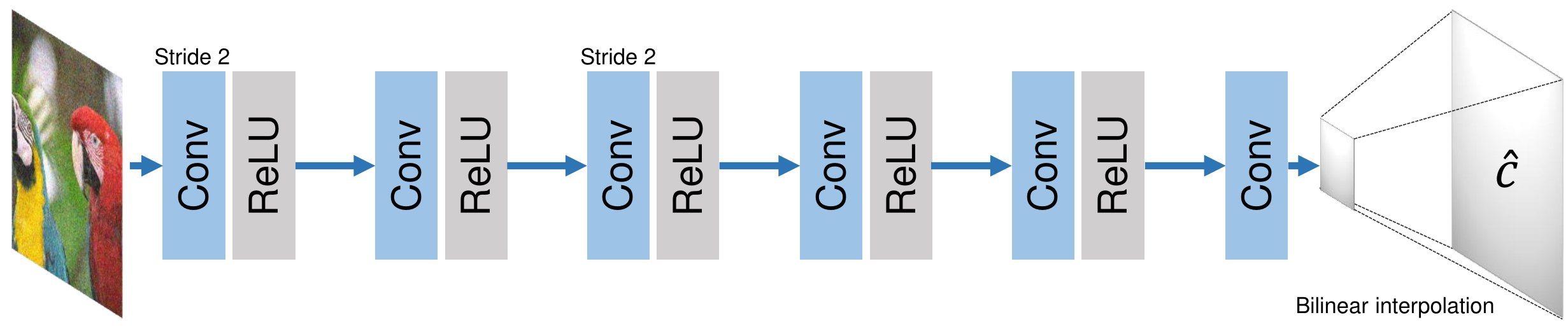}
	\end{center}
	\caption{The network architecture of conditional estimation network (CENet). We adopt $3\times3$ convolution layers for all layers, and the number of channels is $64$ except for the last one. We decrease the spatial size of feature maps with the strided convolution layers. At the last stage, we bilinearly interpolate the map to recover its spatial size back to the same size of the input image.}
	\label{fig:CENet}
\end{figure}

\begin{figure*}[t]
	\begin{center}
		\centering
		\includegraphics[width=0.87\linewidth]{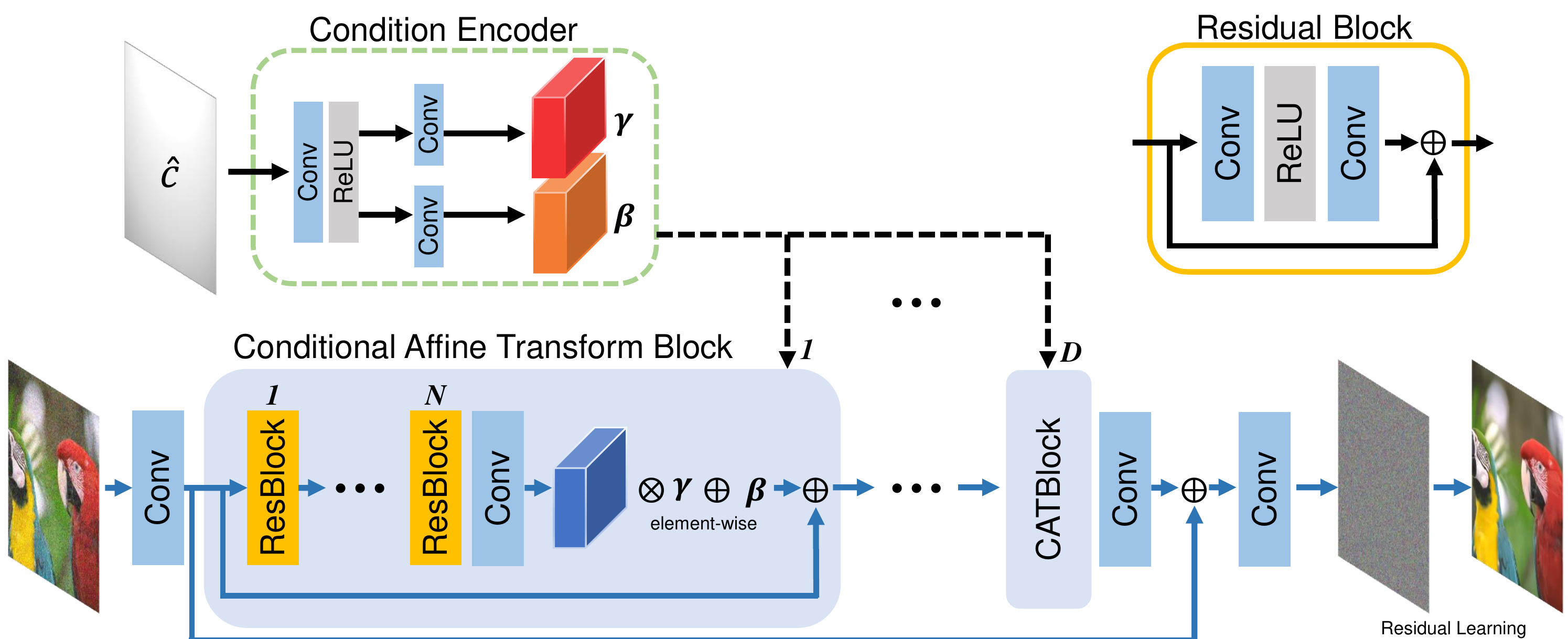}
	\end{center}		
	\caption{The network architecture of tunable denoiser. We adopt $3\times3$ convolution layers for the main stem of the network, including conditional affine transform blocks (CATBlocks) and residual blocks (ResBlocks). Also, the number of channels is $64$ for the main stem, and $1\times1$ convolution layers are employed for the condition encoder. In condition encoder, the number of channels for the first layer is $128$, and the others are $64$.}
	\label{fig:Denoiser}
\end{figure*}

In this section, we introduce the choice of $\c$, and how we design a network for the point estimate $\hatc$. Notably, we consider the cases that $p(\c|\y)$ is unimodal with a sharp peak, for tractable solution, as $\c$ may belong to any category unless it satisfies such properties. Fortunately, a noise-level of a given noisy image is deterministic, which means that its distribution is unimodal with a relatively sharp peak. In other words, only a single noise-level corresponds to the given noisy image. Therefore, we choose the noise-level as $\c$.
In addition to the blind scenario, we can also handle the non-blind case by solving the sub-problem of Eq.~\ref{eq:8} with known information $\c$. Also, we can manually control $\c$ to manipulate the network to produce better outputs.

To estimate the noise-level of a given image with an unknown and spatially varying noise variance, we exploit a CNN architecture shown in \figurename{~\ref{fig:CENet}}. In addition to the choice of $\c$, we add another prior knowledge that the variation of noise-level is spatially smooth in designing the network. We decrease the size of feature maps using the strided convolutions and apply bilinear interpolation at the last stage, rather than using a smoothness term such as total variation regularization. By doing so, we have two advantages compared to the additional loss term. First, we can decrease the computational complexity by decreasing the feature map size, and second, we do not need to tune the hyper-parameter that balances the original loss and the regularization term. In summary, the network is trained with the loss:
\begin{equation}
\mathcal{L}_{c}(\theta)=\mathbb{E} [||\sigma - g_\theta(\y)||^2_2].
\end{equation}
The output of $g_\theta(\y)$ is the estimated noise-level map $\hat{\sigma} \in \mathbb{R}^{H\times W\times C}$, where $H$, $W$, and $C$ denote height, width and number of channels of the noisy image respectively. 

\section{Tunable Denoising Network}
\label{sec:Denoiser}

In this section, we present a novel network architecture where the input is a noisy image with the intervention of an additional tunable parameter $\c$ to emulate $\arg \max p(\x|\y,\hatc)$. The overall architecture of our tunable denoiser is shown in \figurename{~\ref{fig:Denoiser}} where its design directly reflects $f_\phi(y;\hat{c})$. The denoising network mainly consists of two parts: the main stem and condition encoder.

\subsection{Main Stem}
The main stem of our tunable denoising network is based on the cascade of $D$ conditional affine transform blocks (CATBlocks). To incorporate conditional information efficiently, we present a CATBlock which applies the affine transformation to feature maps as
\begin{equation}
F_{o}= \gamma  \otimes F_{i} + \beta,
\end{equation}
where $F_{i}$ and $F_{o}$ denote feature maps before and after the transformation, respectively, and $\otimes$ denotes element-wise multiplication. In the CATBlock, the output feature maps of the last convolution layer are adjusted by affine transformation with respect to the condition-encoded affine parameters $\gamma$ and $\beta$, and the encoded parameters are shared along with other CATBlocks. Similar approaches have been applied in various tasks \cite{AdaIN, FiLM, SFT-GAN, SPADE}.

The CATBlock includes a cascade of $N$ residual blocks (ResBlocks), which is shown in the upper-right corner of \figurename{~\ref{fig:Denoiser}}. Additionally, we adopt the residual learning scheme \cite{DnCNN}, which is to learn noise rather than the clean image itself. Notably, our CATBlocks and ResBlocks adopt residual skip-connection to bypass short- to long-term information.

\subsection{Condition Encoder}
The condition encoder network is a simple two-layers of $1\times1$ convolution. It takes the conditional variable and outputs the condition-encoded parameters $\gamma$ and $\beta$. These parameters adjust the feature values according to the condition. It is notable that the number of parameters of the condition encoder is negligible compared to the whole network.

\subsection{Implementation Details}
Now we specify the implementation details of our DUBD.
For the number of blocks, we set $D=5$ and $N=5$. The DIV2K \cite{DIV2K} dataset is used for the training, which is a recently released high-resolution dataset. It consists of high-quality images of $800$ training, $100$ validation, and $100$ test images. The noise-levels of Gaussian noise are randomly sampled from the range of $[5, 70]$ uniformly. We extract $96\times 96$ size of patches for training. We adopt the ADAM optimizer \cite{ADAM}, and the initial learning rate is set to $2\times10^{-4}$ and halved once during training. Mean squared error (MSE) loss function, which is expressed as
\begin{equation}
\mathcal{L}_{dn}(\phi)=\mathbb{E}[||\x-f_\phi(\y;\c)||^2_2]
\end{equation}
is adopted for denoising loss.

\begin{table*}[!t]
	\caption{The average PSNR results on test sets. The best results are denoted in \textcolor{red}{red} and the second best in \textcolor{blue}{blue}. `*' next to the name of the method means blind denoising.}
	\begin{center}
		\resizebox{1.0\linewidth}{!}{
			\begin{tabular}{|c|c|c|c|c|c|c|c|c|c|c|c|c|}
				\hline
				\rule[-1ex]{0pt}{3.5ex}
				Noise-level & Dataset & CBM3D \cite{BM3D} & TNRD \cite{TNRD} & RED \cite{RED} & MemNet* \cite{MemNet} & CDnCNN* \cite{DnCNN} & FFDNet \cite{FFDNet} & UNLNet* \cite{UNLNet} & ATDNet* \cite{ATDNet} & VDN* \cite{VDN} & DUBD-NB (Ours) & DUBD-B* (Ours)\\
				\hline\hline
				
				\rule[-1ex]{0pt}{3.5ex}
				\multirow{3}{*}{$\sigma = 10$}&CBSD68&35.91&-&33.89&28.52&36.13&36.14& 36.20 & 36.29 & 36.29 & \textcolor{red}{36.35} & \textcolor{blue}{36.33}\\
				\rule[-1ex]{0pt}{3.5ex}
				&Kodak24  & 36.43 & - & 34.73 & 29.70 & 36.46 & 36.69 & - &36.98 & 36.85 & \textcolor{red}{37.03} & \textcolor{blue}{37.02}\\
				\rule[-1ex]{0pt}{3.5ex}
				&Urban100& 36.00 & - & 34.42 & 29.44 & 34.61 & 35.78 & - & \textcolor{blue}{36.31} & 35.97 & \textcolor{red}{36.32} & 36.23 \\
				\hline\hline
				
				\rule[-1ex]{0pt}{3.5ex}
				\multirow{3}{*}{$\sigma = 30$}& CBSD68 & 29.73 & - & 28.45 & 28.39 & 30.34 & 30.32 & 30.21 & 30.61 & \textcolor{blue}{30.64} & \textcolor{red}{30.65} & 30.62 \\
				\rule[-1ex]{0pt}{3.5ex}
				&Kodak24  & 30.75 & -& 29.53 & 29.55 & 31.17 & 31.27 & 31.18 & \textcolor{blue}{31.72} & 31.67 & \textcolor{red}{31.75} & \textcolor{red}{31.75} \\
				\rule[-1ex]{0pt}{3.5ex}
				&Urban100& 30.36&-&28.84 & 28.93 & 30.00 & 30.53 & 30.41 & \textcolor{red}{31.48} & 31.14 & \textcolor{blue}{31.46} & 31.43 \\
				\hline\hline
				
				\rule[-1ex]{0pt}{3.5ex}
				\multirow{3}{*}{$\sigma = 50$}&CBSD68& 27.38& 25.96 & 26.34 & 26.33 & 27.95 & 27.97 & 27.85 & \textcolor{blue}{28.33} & \textcolor{blue}{28.33} & \textcolor{red}{28.35} & 28.31 \\
				\rule[-1ex]{0pt}{3.5ex}
				&Kodak24  & 28.46 & 27.04 & 27.42  & 27.51 & 28.83 & 28.98 & 28.86 &29.48 & 29.44 &\textcolor{red}{29.51} & \textcolor{blue}{29.50} \\
				\rule[-1ex]{0pt}{3.5ex}
				&Urban100& 27.94 & 25.52 & 26.25 & 26.53 & 27.59 & 28.05 & 27.95 &\textcolor{red}{29.20} & 28.86 & \textcolor{blue}{29.16} & 29.14 \\
				\hline\hline
				
				\rule[-1ex]{0pt}{3.5ex}
				\multirow{3}{*}{$\sigma = 70$}&CBSD68& 26.00 & - & 25.09 & 25.09 & 25.66 & 26.55 & - & - & \textcolor{blue}{26.93}& \textcolor{red}{26.96} & 26.89 \\
				\rule[-1ex]{0pt}{3.5ex}
				&Kodak24  & 27.09 & - & 26.16 & 26.24 & 26.36 & 27.56 & - & - & 28.05 &\textcolor{red}{28.12} & \textcolor{blue}{28.11} \\
				\rule[-1ex]{0pt}{3.5ex}
				&Urban100& 26.31 & - & 24.58 & 24.93 & 25.24 & 26.40 & - & - & 27.31 &\textcolor{red}{27.59} & \textcolor{blue}{27.58} \\
				
				\hline
			\end{tabular}
		}
	\end{center}
	
	\label{table:results}
\end{table*}

\begin{table*}[!t]
	\caption{The number of parameters and PSNR for several CNN-based methods. The PSNR is measured on Urban100 \cite{Urban} set with $\sigma=50$. For RNAN \cite{RNAN}, we brought the number from the paper.}
	\begin{center}
		\resizebox{0.9\linewidth}{!}{
			\begin{tabular}{|c|c|c|c|c|c|c|c|c|}
				\hline
				\rule[-1ex]{0pt}{3.5ex}
				Methods & RED \cite{RED} & CDnCNN* \cite{DnCNN} & FFDNet \cite{FFDNet} &  ATDNet* \cite{ATDNet} & RNAN \cite{RNAN} & VDN* \cite{VDN} & DUBD-NB (Ours) & DUBD-B* (Ours) \\
				\hline\hline
				\rule[-1ex]{0pt}{3.5ex}
				Parameters &  4,135 K & 668 K & 825 K & 9,453 K & 7,409 K & 7,817 K & 2,088 K & 2,239 K \\
				\hline
				\rule[-1ex]{0pt}{3.5ex}
				PSNR (dB) & 26.25 & 27.59 & 28.05 & 29.20 & 29.08 & 28.86 & 29.16 & 29.14\\
				\hline
				
			\end{tabular}
		}
	\end{center}
	
	\label{table:params}
\end{table*}

\begin{figure*}[!t]
	\begin{center}
		\begin{subfigure}[t]{0.24\linewidth}
			\centering
			\includegraphics[width=1\columnwidth]{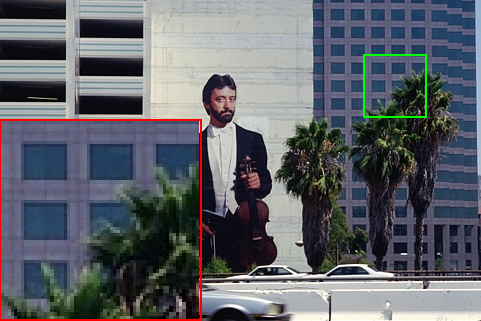}
			\caption*{Ground Truth}
		\end{subfigure}
		\begin{subfigure}[t]{0.24\linewidth}
			\centering
			\includegraphics[width=1\columnwidth]{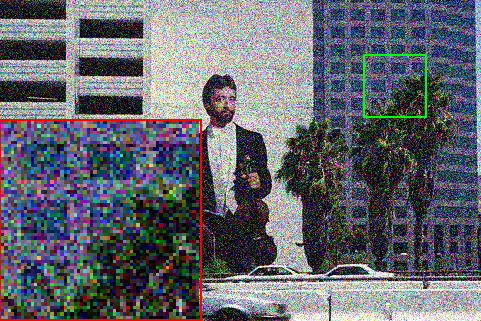}
			\caption*{Noisy}
		\end{subfigure}
		\begin{subfigure}[t]{0.24\linewidth}
			\centering
			\includegraphics[width=1\columnwidth]{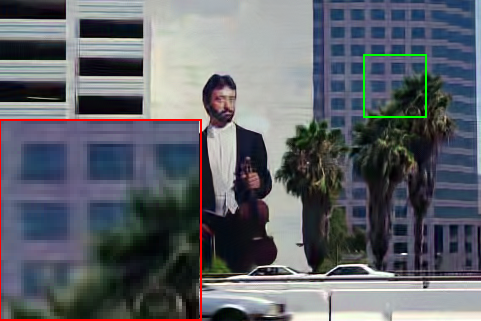}
			\caption*{CBM3D \cite{BM3D}}
		\end{subfigure}
		\begin{subfigure}[t]{0.24\linewidth}
			\centering
			\includegraphics[width=1\columnwidth]{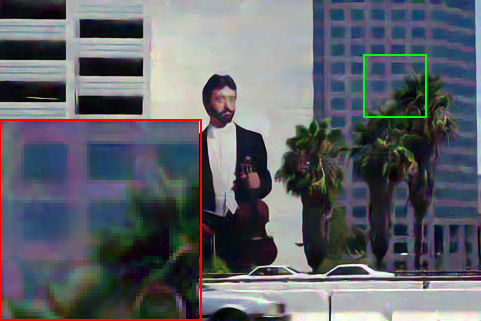}
			\caption*{TNRD \cite{TNRD}}
		\end{subfigure}
		\begin{subfigure}[t]{0.24\linewidth}
			\centering
			\includegraphics[width=1\columnwidth]{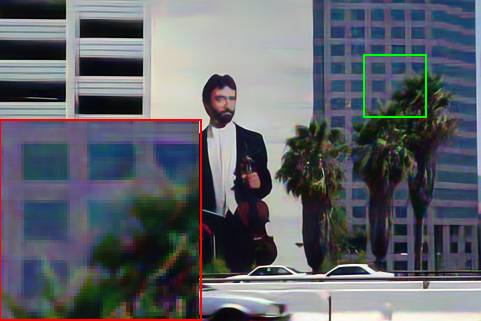}
			\caption*{RED \cite{RED}}
		\end{subfigure}
		\begin{subfigure}[t]{0.24\linewidth}
			\centering
			\includegraphics[width=1\columnwidth]{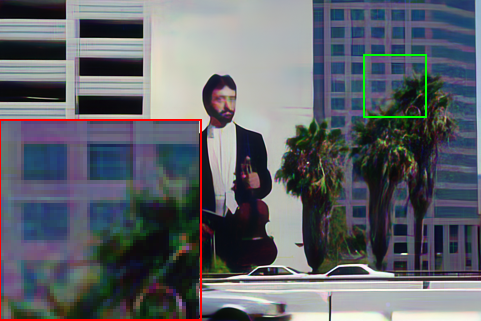}
			\caption*{MemNet \cite{MemNet}}
		\end{subfigure}
		\begin{subfigure}[t]{0.24\linewidth}
			\centering
			\includegraphics[width=1\columnwidth]{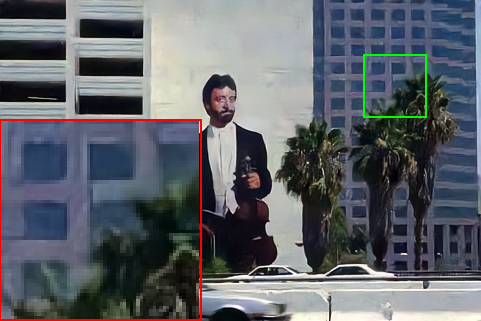}
			\caption*{CDnCNN \cite{DnCNN}}
		\end{subfigure}
		\begin{subfigure}[t]{0.24\linewidth}
			\centering
			\includegraphics[width=1\columnwidth]{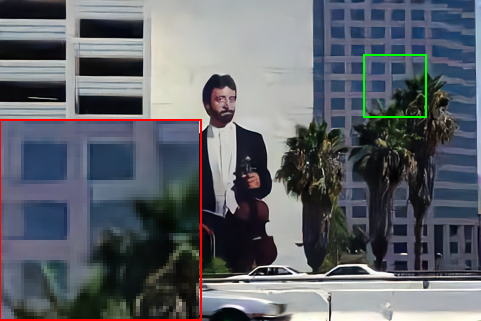}
			\caption*{FFDNet \cite{FFDNet}}
		\end{subfigure}
		\begin{subfigure}[t]{0.24\linewidth}
			\centering
			\includegraphics[width=1\columnwidth]{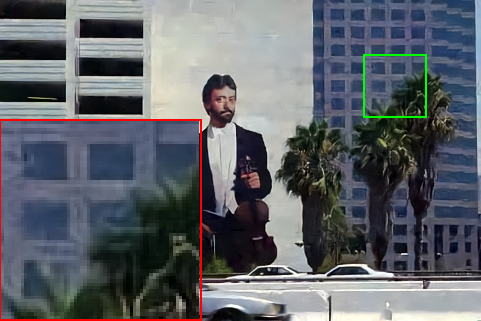}
			\caption*{UNLNet \cite{UNLNet}}
		\end{subfigure}
		\begin{subfigure}[t]{0.24\linewidth}
			\centering
			\includegraphics[width=1\columnwidth]{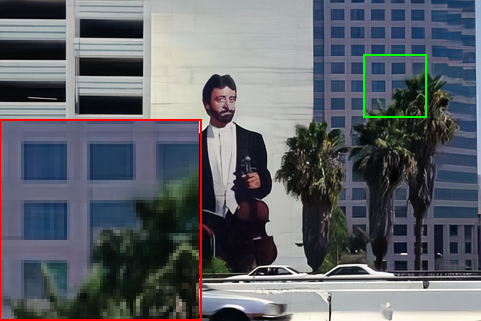}
			\caption*{ATDNet \cite{ATDNet}}
		\end{subfigure}
		\begin{subfigure}[t]{0.24\linewidth}
			\centering
			\includegraphics[width=1\columnwidth]{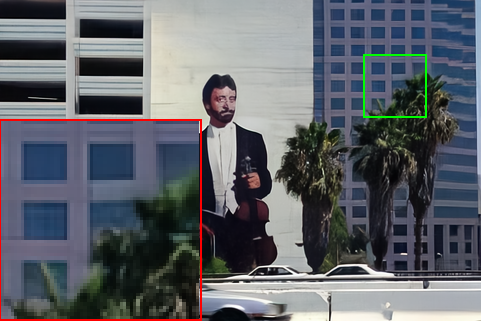}
			\caption*{DUBD-NB (Ours)}
		\end{subfigure}
		\begin{subfigure}[t]{0.24\linewidth}
			\centering
			\includegraphics[width=1\columnwidth]{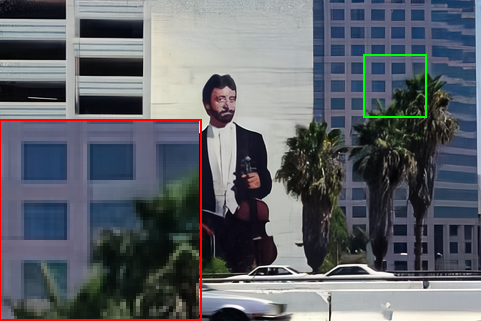}
			\caption*{DUBD-B (Ours)}
		\end{subfigure}

	\end{center}
	\caption{Visualized examples of denoising for $\sigma=50$.}
	\label{fig:001}
\end{figure*}

\begin{figure*}[!t]
	\begin{center}
		\begin{subfigure}[t]{0.16\linewidth}
			\centering
			\includegraphics[width=1\columnwidth]{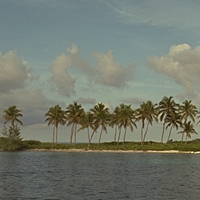}
			\caption*{Ground Truth}
		\end{subfigure}
		\begin{subfigure}[t]{0.16\linewidth}
			\centering
			\includegraphics[width=1\columnwidth]{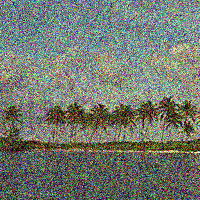}
			\caption*{Noisy}
		\end{subfigure}
		\begin{subfigure}[t]{0.16\linewidth}
			\centering
			\includegraphics[width=1\columnwidth]{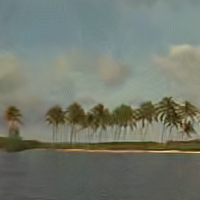}
			\caption*{CBM3D \cite{BM3D}}
		\end{subfigure}
		\begin{subfigure}[t]{0.16\linewidth}
			\centering
			\includegraphics[width=1\columnwidth]{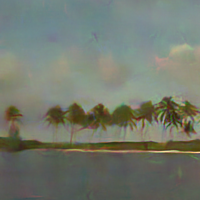}
			\caption*{TNRD \cite{TNRD}}
		\end{subfigure}
		\begin{subfigure}[t]{0.16\linewidth}
			\centering
			\includegraphics[width=1\columnwidth]{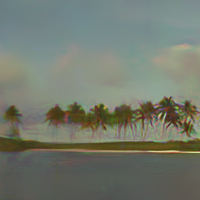}
			\caption*{RED \cite{RED}}
		\end{subfigure}
		\begin{subfigure}[t]{0.16\linewidth}
			\centering
			\includegraphics[width=1\columnwidth]{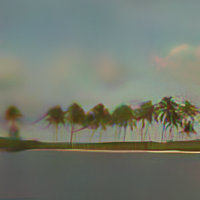}
			\caption*{MemNet \cite{MemNet}}
		\end{subfigure}
		\begin{subfigure}[t]{0.16\linewidth}
			\centering
			\includegraphics[width=1\columnwidth]{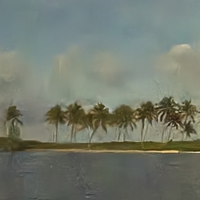}
			\caption*{CDnCNN \cite{DnCNN}}
		\end{subfigure}
		\begin{subfigure}[t]{0.16\linewidth}
			\centering
			\includegraphics[width=1\columnwidth]{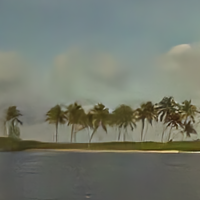}
			\caption*{FFDNet \cite{FFDNet}}
		\end{subfigure}
		\begin{subfigure}[t]{0.16\linewidth}
			\centering
			\includegraphics[width=1\columnwidth]{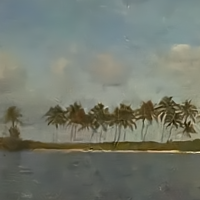}
			\caption*{UNLNet \cite{UNLNet}}
		\end{subfigure}
		\begin{subfigure}[t]{0.16\linewidth}
			\centering
			\includegraphics[width=1\columnwidth]{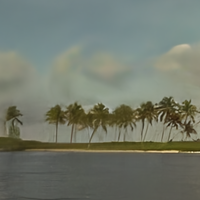}
			\caption*{ATDNet \cite{ATDNet}}
		\end{subfigure}
		\begin{subfigure}[t]{0.16\linewidth}
			\centering
			\includegraphics[width=1\columnwidth]{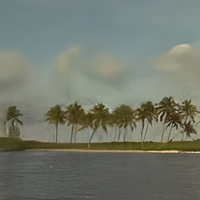}
			\caption*{DUBD-NB (Ours)}
		\end{subfigure}
		\begin{subfigure}[t]{0.16\linewidth}
			\centering
			\includegraphics[width=1\columnwidth]{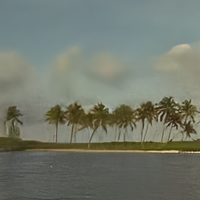}
			\caption*{DUBD-B (Ours)}
		\end{subfigure}

	\end{center}
	\caption{Visualized examples of denoising for $\sigma=50$.}
	\label{fig:002}
\end{figure*}

\section{Experimental Results}
\label{sec:experiments}

We test denoising performances on the images corrupted with noise-levels $\sigma=10,~30,~50,~70$, on three famous denoising datasets for color images: CBSD68 \cite{BSD}, Kodak24, and Urban100 \cite{Urban}.
Importantly, considering the practical usage, we mainly aim color image denoising where gray image denoising can also be done with the change of input channels. 

We compare our method with several denoising algorithms from conventional non-learning methods to recent CNN-based state-of-the-arts: CBM3D \cite{BM3D}, TNRD \cite{TNRD}, RED \cite{RED}, MemNet \cite{MemNet}, DnCNN \cite{DnCNN}, FFDNet \cite{FFDNet}, UNLNet \cite{UNLNet}, and ATDNet \cite{ATDNet}. We present two results of our DUBD, namely DUBD-NB and DUBD-B. The DUBD-NB is a non-blind model where we assume known noise-levels, and thus the output corresponds to $f_\phi (\y;\c)$. On the other hand, DUBD-B is a blind model where the denoiser takes the estimated $\c$ from CENet, and hence the output is $f_\phi (\y;\hatc)$. It is notable that we train only \emph{one} universal network for our DUBD that can be used both as blind and non-blind, whereas other methods include blind and non-blind models separately for each purpose. We evaluate the results by PSNR (peak signal-to-noise ratio) on color images.{\tiny }

The overall results are listed in \tablename{~\ref{table:results}}, which shows that our methods are on par with or exceeding previous state-of-the-art methods. Notably, our DUBD-B shows much better performances than other non-blind ones. Interestingly, DUBD-B and DUBD-NB show negligible performance gaps, even though the estimation is not quite perfect. In other words, our method is robust or not very sensitive to the conditional estimations. ATDNet \cite{ATDNet} blind model shows comparable results to ours or sometimes shows the best results among the others. But the applicable range of noise-level of ATDNet is narrower than ours. Also, an interesting fact is that CBM3D \cite{BM3D}, which is a conventional approach exploiting non-local self-similarity with 3D transformed Wiener filter, shows quite great results better than most of the CNN-based methods in Urban100 \cite{Urban} set that have many recurrent patterns.

For visual comparisons, we show some denoising results in \figurename{~\ref{fig:001}} and \ref{fig:002}. As shown, ours show a better separation of content and noise, \emph{i.e}, noise reduction. For other methods, as shown in \figurename{~\ref{fig:002}}, they often show cloudy patterns or entangled noise, and sometimes show color artifacts. On the other hand, our DUBDs achieve plausible detail preservation and naturalness.

\section{Analysis}

We further analyze our DUBD with other recent state-of-the-art methods, focusing on the complexity and robustness to spatially or spectrally varying noises, which are important factors for practicability. 

\begin{figure}[t]
	\begin{center}
		\centering
		\includegraphics[width=0.85\linewidth]{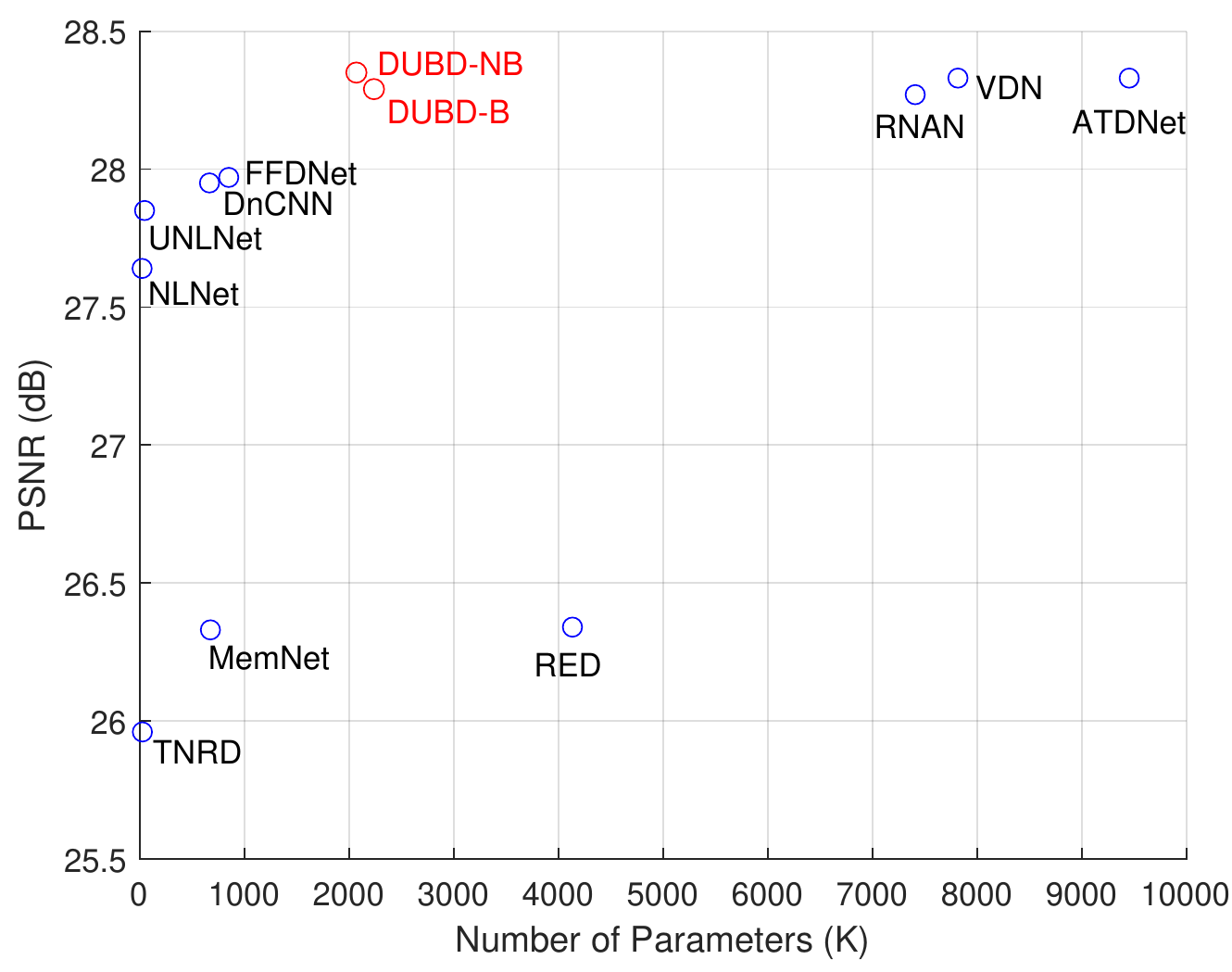}
	\end{center}		
	\caption{The number of parameter vs. PSNR on CBSD68 with $\sigma =50$.}
	\label{fig:params}
\end{figure}

\subsection{The Number of Parameters}

Since the computational resource is limited in a real-world situation, the number of parameters is an important factor for the practicability of CNNs. \tablename{~\ref{table:params}} and \figurename~\ref{fig:params} show the summary of network complexity vs. performance for the methods compared earlier. Compared to CDnCNN \cite{DnCNN} and FFDNet \cite{FFDNet}, our methods show PSNR gain more than $1.0$ to $1.5$ dB by doubling or tripling the number of parameters. Our DUBD outperforms RED \cite{RED} with about half the number of parameters. Also, compared to ATDNet \cite{ATDNet} and RNAN \cite{RNAN}, our methods need much fewer parameters, while showing comparable PSNRs. Importantly, RNAN \cite{RNAN} is a non-blind method which requires each separate model for a specific noise-level. Additionally, in our methods, changing from non-blind method to blind method requires only 151 K parameters for CENet. We also visualize the comparisons of several methods in terms of parameters versus performance in \figurename{~\ref{fig:params}}. It is noteworthy that although we both include blind and non-blind methods in the comparisons, our DUBD shows outstanding performance with a comparable number of parameters. In summary, our DUBD is practical and competitive compared to the state-of-the-art methods.

\subsection{Dealing with Spectrally-Spatially Variant Noise}

In a real-world situation, the noise in an image is rarely spatially uniform. Sometimes, the noise-level varies in different color channels, as pointed out in \cite{FFDNet, TWSC}. Thus, in this section, we show the results of spectrally-spatially varying noise with our DUBD-B.

\subsubsection{Spectrally Variant Noise}
\begin{figure}[t]
	\begin{center}
		\begin{subfigure}[t]{0.32\linewidth}
			\centering
			\includegraphics[width=1\columnwidth]{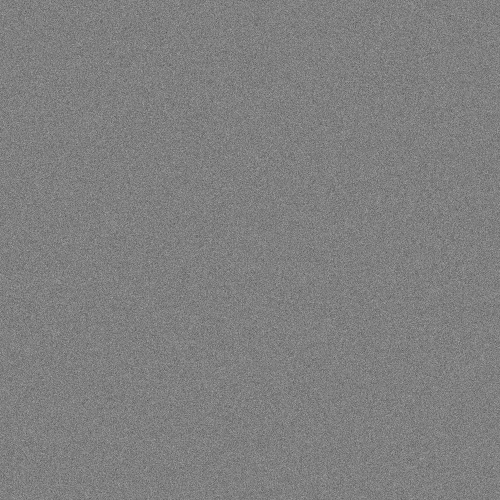}
			\caption{Noise in R}
		\end{subfigure}
		\begin{subfigure}[t]{0.32\linewidth}
			\centering
			\includegraphics[width=1\columnwidth]{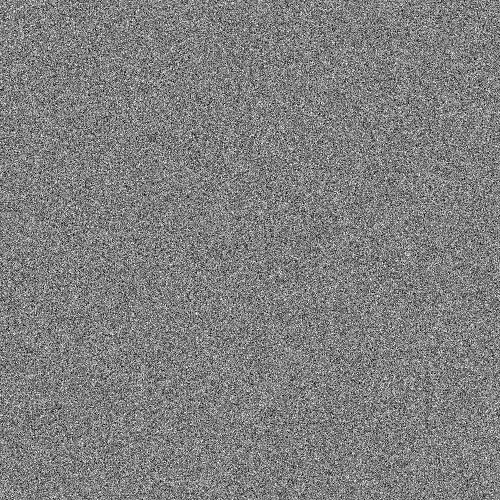}
			\caption{Noise in G}
		\end{subfigure}
		\begin{subfigure}[t]{0.32\linewidth}
			\centering
			\includegraphics[width=1\columnwidth]{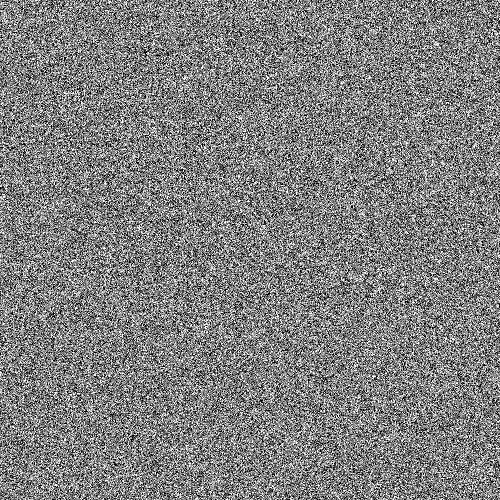}
			\caption{Noise in B}
		\end{subfigure}
		\begin{subfigure}[t]{0.32\linewidth}
			\centering
			\includegraphics[width=1\columnwidth]{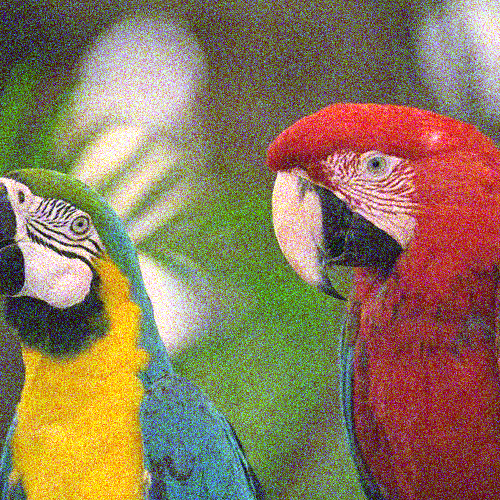}
			\caption{Input noisy image}
		\end{subfigure}
		\begin{subfigure}[t]{0.32\linewidth}
			\centering
			\includegraphics[width=1\columnwidth]{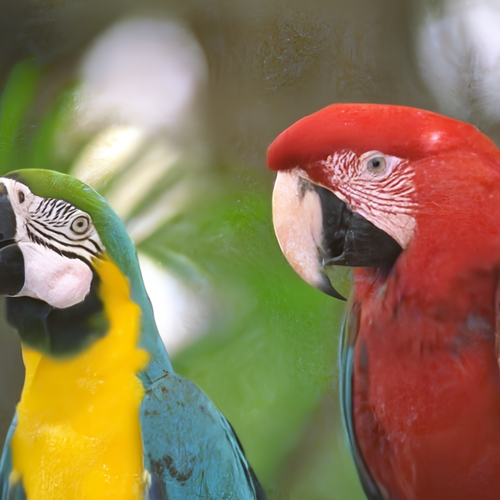}
			\caption{Output with average noise-level}
		\end{subfigure}
		\begin{subfigure}[t]{0.32\linewidth}
			\centering
			\includegraphics[width=1\columnwidth]{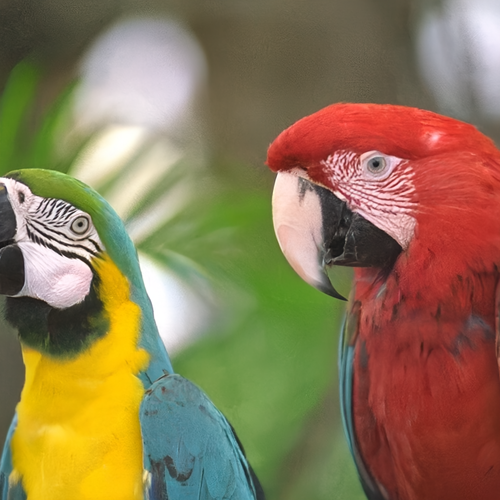}
			\caption{Our result}
		\end{subfigure}
	\end{center}
	\caption{Input and results on spectrally varying noise.}
	\label{fig:spectral}
\end{figure}

To test on spectrally variant noise, we feed an input as shown in \figurename~\ref{fig:spectral}(d), where the extent of noise corruption differs in each color channel. As shown in \figurename{~\ref{fig:spectral}}(a-c), the noise corruption is severe in B-channel while it is mild in R-channel. As shown in \figurename~\ref{fig:spectral}(e), addressing this type of noise with only a single averaged noise-level does not show plausible results. It can be seen that the noise in B-channel (appears in the bluish region of the image) remains after the denoising, while there is loss-of-details of the red bird. On the other hand, since our method can handle each channel separately, it shows visually promising results as in \figurename~\ref{fig:spectral}(f).

\subsubsection{Spatially Variant Noise}

\begin{figure}[t]
	\begin{center}
		\begin{subfigure}[t]{0.35\linewidth}
			\centering
			\includegraphics[width=1\columnwidth]{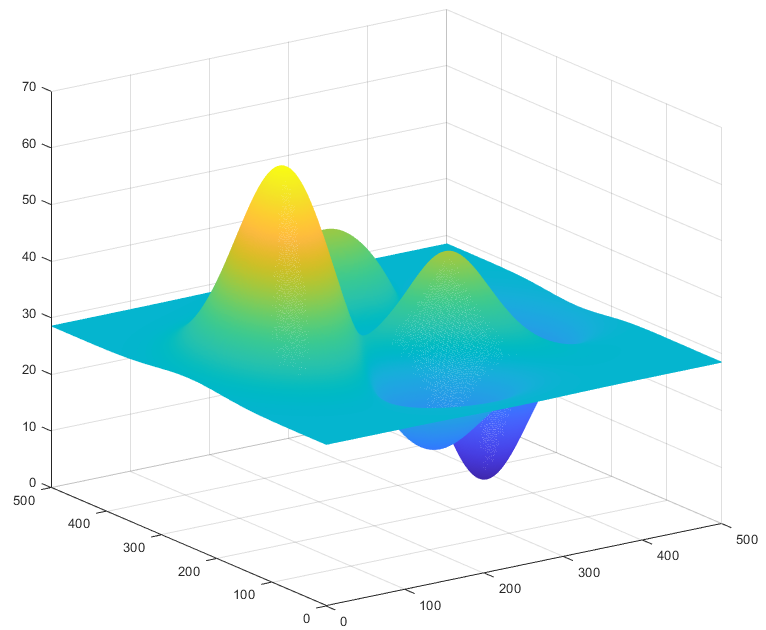}
			\caption{Noise-level in spatial dimension}
		\end{subfigure}
		\begin{subfigure}[t]{0.35\linewidth}
			\centering
			\includegraphics[width=1\columnwidth]{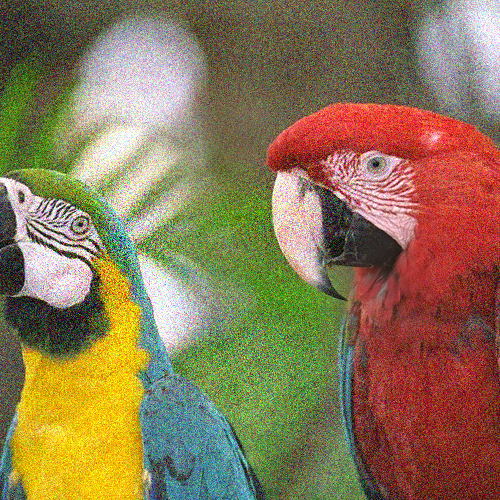}
			\caption{Input noisy image}
		\end{subfigure}
		\begin{subfigure}[t]{0.35\linewidth}
			\centering
			\includegraphics[width=1\columnwidth]{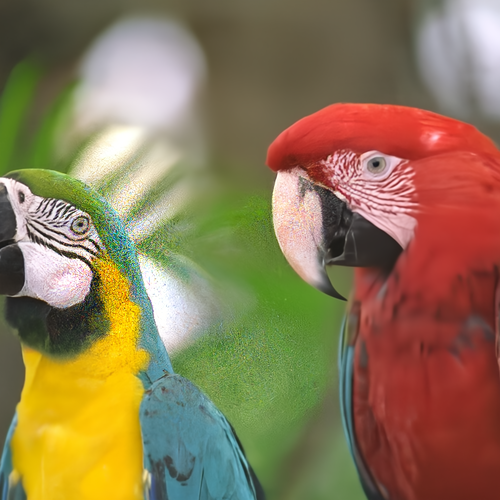}
			\caption{Output with average noise-level}
		\end{subfigure}
		\begin{subfigure}[t]{0.35\linewidth}
			\centering
			\includegraphics[width=1\columnwidth]{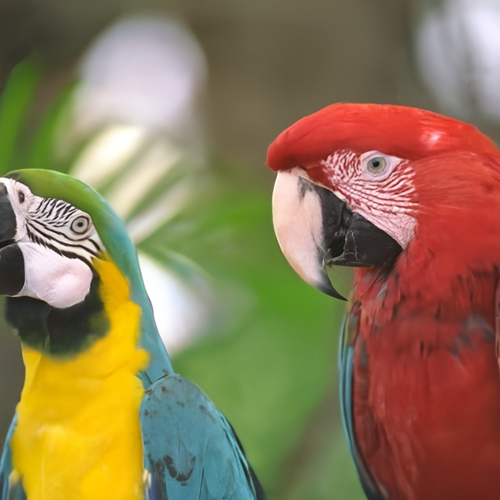}
			\caption{Our result}
		\end{subfigure}
	\end{center}
	\caption{Input and results on spatially varying noise.}
	\label{fig:spatial}
\end{figure}

We feed an input image with spatially variant noise as shown in \figurename{~\ref{fig:spatial}}(b). The noise-level of each spatial position is shown in \figurename{~\ref{fig:spatial}}(a) and the results are shown in \figurename{~\ref{fig:spatial}}(c) and (d). As shown in the results, denoising with a single noise-level shows over-smoothed regions, as in the red bird, and there are some regions that the noise is not removed yet. On the other hand, our method handles spatially variant noise well within the blind denoising scheme, showing visually pleasing results.

\subsection{Traversing Conditional Variable}

\begin{figure*}[t]
	\begin{center}
		\begin{subfigure}[t]{0.20\linewidth}
			\centering
			\includegraphics[width=1\columnwidth]{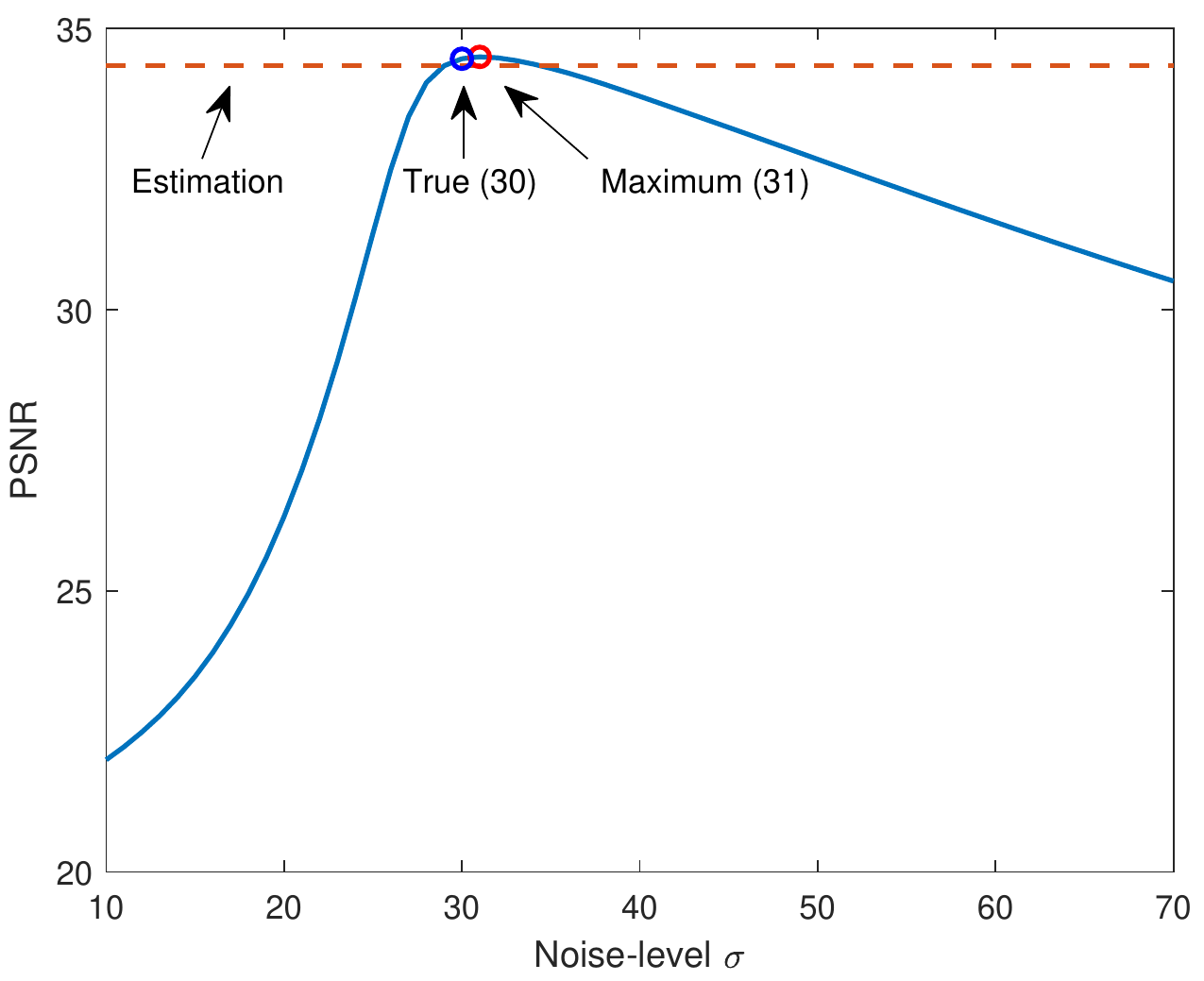}
			\caption{PSNR results on different input $\c$.}
		\end{subfigure}
		\begin{subfigure}[t]{0.78\linewidth}
			\centering
			\includegraphics[width=1\columnwidth]{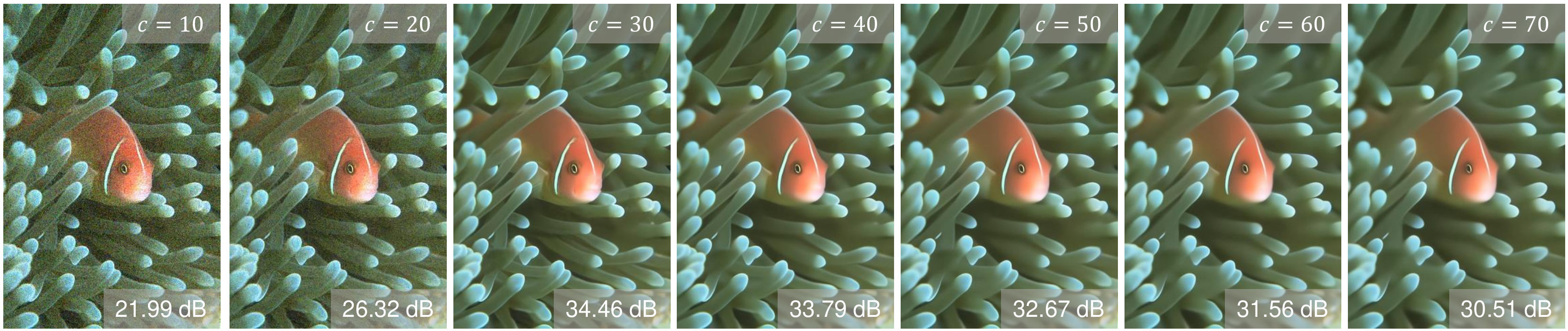}
			\caption{Visualized results of denoised images with different input $\c$.}
		\end{subfigure}
	\end{center}
	\caption{Results by traversing the conditional variable $\c$.}
	\label{fig:traverse}
\end{figure*}

In this section, we show the results by traversing the conditional variable $\c$ for two reasons. First, we show that our tunable denoiser can also be manually controlled in accordance with user preferences. By controlling the $\c$, the denoising strength can be tuned. Second, we attempt to investigate the effect of conditional variable and indirectly trace the conditional posterior $p(\c|\y)$.

The results are shown in \figurename{~\ref{fig:traverse}}, where we feed an image corrupted by i.i.d. Gaussian noise with $\sigma=30$ by changing the input $\c$ of our tunable denoiser. As shown in the \figurename{~\ref{fig:traverse}}(a), the PSNR graph achieves its peak at $\c=31$, near the true $\sigma$. Also, our blind result with conditional estimation shows quite a competent result. Notably, the graph shape is asymmetric unimodal, where higher $\c$ than the true value shows much less sensitivity then feeding the lower $\c$ than the true value. Its coherent tendency can also be found in the corresponding visualizations on \figurename{~\ref{fig:traverse}(b)}. Starting from small $\c$, the noise remains in the image, but by tuning $\c$ to higher values, the noise gradually diminishes. Then, increasing $\c$ higher than the true noise-level, the image gets blurry, losing details and textures.

\section{Results on Real Noisy Images}

As the noise from the real-world largely deviates from the AWGN, it is important to validate whether the network will also work well for the real-world noises, with possible modification or retraining. Real-noise originates from the shot-noise and read-noise from the sensors and then spreads throughout the camera pipeline. Hence, the noise is not i.i.d. any more, and it is signal-dependent and correlated with neighboring pixels due to demosaicking. Hence, we also train our network to deal with real-noise images. To train our network for real image denoising, we use Smartphone Image Denoising Dataset (SIDD) \cite{SIDD}, which includes pairs of clean and noisy images from smartphone cameras. For the choice of $\c$, since we do not know the noise-level, we use the average pixel value of the noisy image of $4\times4$ region as our $\c$. It can be expressed as
\begin{equation}
\label{eq:real}
\c=Avgpool_{4\times 4} (\y).
\end{equation}
Then, the CENet becomes just a simple average pooling operation, and our tunable denoiser targets different color valued regions separately. That is, it can be operated within the divide-and-conquer scheme. We choose a widely-used benchmark for the evaluation: Darmstadt Noise Dataset (DND) \cite{DND}, which consists of 50 images with real-noise from 50 scenes and the scenes are further cropped by the provider which results in 1,000 small patches. The ground-truth image is not available, and the evaluation can only be done through online submission.

\subsection{Analysis}
We first analyze the effect of the choice of $\c$. As our ablation investigation, we train a baseline DUBD model within an end-to-end scheme without any constraints on the conditional variable $\c$. Then, we trained our DUBD-R (R stands for the model for real noise) by using the spatial average color value as our conditional variable. \tablename{~\ref{table:real_ablation} shows the summary. It is shown that by setting $\c$ as Eq. \ref{eq:real}, the PSNR has been improved compared to the baseline model. Also, we note that DUBD-R has a smaller number of parameters compared to the baseline model by replacing the CENet by a simple average pooling method.
		
\begin{table}[!t]
	\caption{Ablation investigation for the real-noise denoising.}
	\begin{center}
		\resizebox{0.6\linewidth}{!}{
			\begin{tabular}{|c|c|c|}
				\hline
				\rule[-1ex]{0pt}{3.5ex}
				Dataset & SIDD \cite{SIDD} & DND \cite{DND} \\
				\hline\hline
				\rule[-1ex]{0pt}{3.5ex}
				Baseline & 39.13 & 39.13 \\
				\hline
				\rule[-1ex]{0pt}{3.5ex}
				DUBD-R & 39.27 & 39.38 \\
				\hline
				
			\end{tabular}
		}
	\end{center}
	
	\label{table:real_ablation}
\end{table}

\subsection{Results}

\begin{table}[!t]
	\caption{Average PSNR and SSIM results on DND \cite{DND} benchmark. The best results are in \textcolor{red}{red} and the second bests are in \textcolor{blue}{blue}.}
	\begin{center}
		\begin{tabular}{|l|c|c|c|}
			\hline
			\rule[-1ex]{0pt}{3.5ex}
			Method & Parameters & PSNR & SSIM \\
			\hline\hline
			\rule[-1ex]{0pt}{3.5ex}
			DnCNN+ \cite{DnCNN} & 668 K & 37.90 & 0.9430 \\
			\rule[-1ex]{0pt}{3.5ex}
			FFDNet+ \cite{FFDNet}& 825 K & 37.61 & 0.9415 \\
			\rule[-1ex]{0pt}{3.5ex}
			CBDNet \cite{CBDNet} & 4,365 K & 38.06 & 0.9421 \\
			\rule[-1ex]{0pt}{3.5ex}
			ATDNet \cite{ATDNet} & 9,453 K & 39.19 & 0.9526 \\
			\rule[-1ex]{0pt}{3.5ex}
			RIDNet \cite{RIDNet} & 1,499 K & 39.26 & \textcolor{blue}{0.9528} \\
			\rule[-1ex]{0pt}{3.5ex}
			VDN \cite{VDN} & 7,817 K & \textcolor{blue}{39.38} & 0.9518 \\
			\rule[-1ex]{0pt}{3.5ex}
			DUBD-R (Ours)& 2,088 K & 39.38 & 0.9526 \\
			\rule[-1ex]{0pt}{3.5ex}
			DUBD-R+ (Ours) & 2,088 K & \textcolor{red}{39.44} & \textcolor{red}{0.9530} \\
			
			\hline
			
		\end{tabular}
	\end{center}	
	\label{table:real_result}
\end{table}

In \tablename{~\ref{table:real_result}}, we demonstrate several comparisons on two real-noise image benchmark datasets. The comparisons are DnCNN+ \cite{DnCNN}, FFDNet+ \cite{FFDNet}, CBDNet \cite{CBDNet}, ATDNet \cite{ATDNet}, RIDNet \cite{RIDNet}, and VDN \cite{VDN}. To achieve further performance gain, we adopt a self-ensemble approach to our method and denote it as ``DUBD-R+.'' As shown in the table, our method achieves competent results. Compared to a recently published blind denoiser VDN \cite{VDN}, our method shows competent results while requiring a much smaller number of parameters. We also visualize comparisons on \figurename{~\ref{fig:real}} for qualitative evaluation. We expect further investigation on the choice of $\c$ can boost the performance for real-world noisy images.

\begin{figure}[t]
	\begin{center}
		\captionsetup{justification=centering}
		\begin{subfigure}[t]{0.24\linewidth}
			\centering
			\includegraphics[width=1\columnwidth]{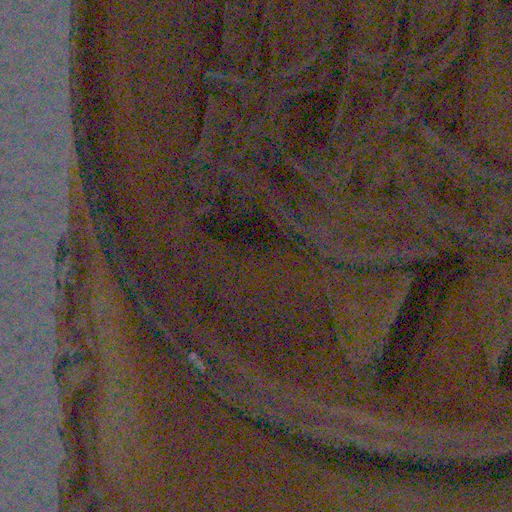}
			\caption*{Noisy \\ 23.55/0.5185}
		\end{subfigure}			
		\begin{subfigure}[t]{0.24\linewidth}
			\centering
			\includegraphics[width=1\columnwidth]{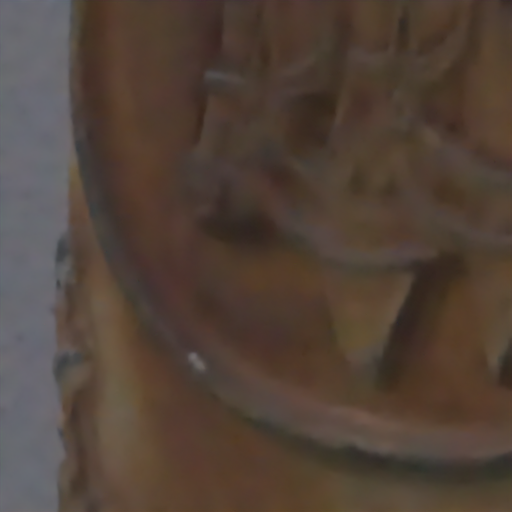}
			\caption*{DnCNN+ \cite{DnCNN} \\ 34.51/0.9457}
		\end{subfigure}
		\begin{subfigure}[t]{0.24\linewidth}
			\centering
			\includegraphics[width=1\columnwidth]{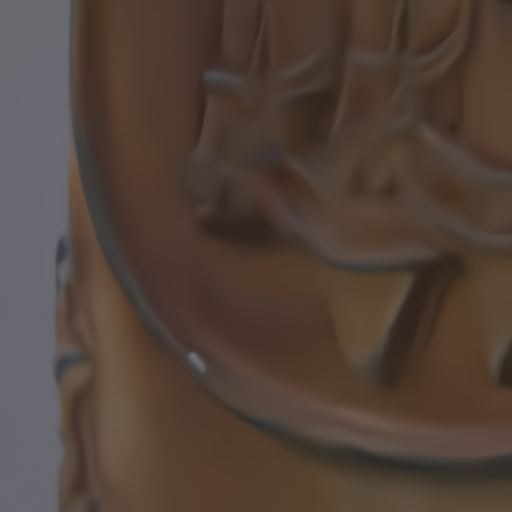}
			\caption*{FFDNet+ \cite{FFDNet} \\ 34.47/0.9510}
		\end{subfigure}			
		\begin{subfigure}[t]{0.24\linewidth}
			\centering
			\includegraphics[width=1\columnwidth]{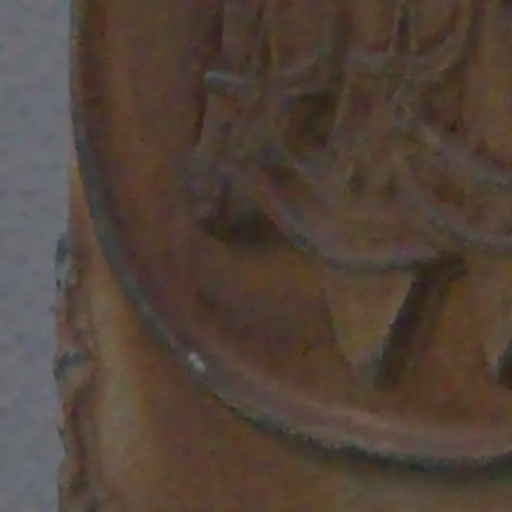}
			\caption*{CBDNet \cite{CBDNet} \\ 35.43/0.9469}
		\end{subfigure}
	
		\begin{subfigure}[t]{0.24\linewidth}
			\centering
			\includegraphics[width=1\columnwidth]{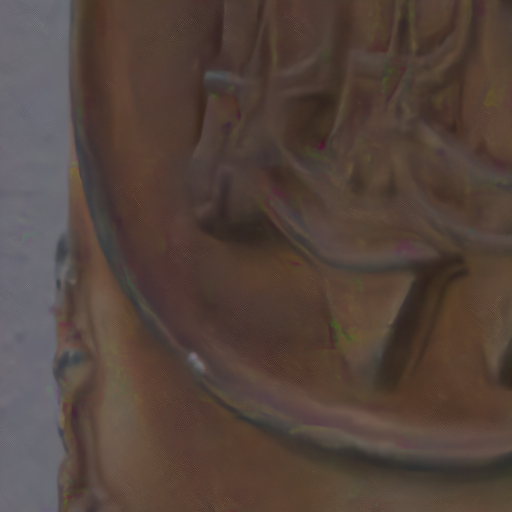}
			\caption*{ATDNet \cite{ATDNet} \\ 36.03/0.9506}
		\end{subfigure}			
		\begin{subfigure}[t]{0.24\linewidth}
			\centering
			\includegraphics[width=1\columnwidth]{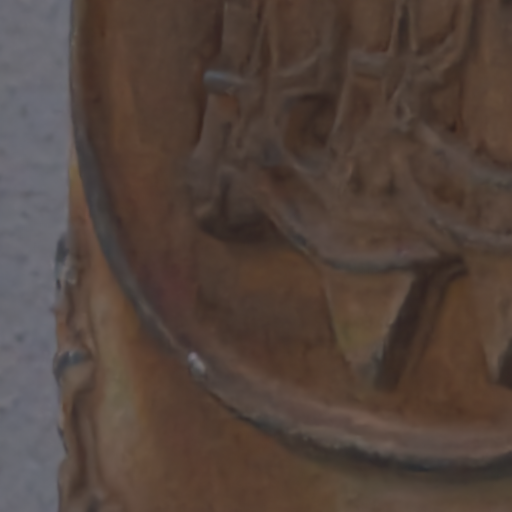}
			\caption*{RIDNet \cite{RIDNet} \\ 37.17/0.9596}
		\end{subfigure}
				\begin{subfigure}[t]{0.24\linewidth}
			\centering
			\includegraphics[width=1\columnwidth]{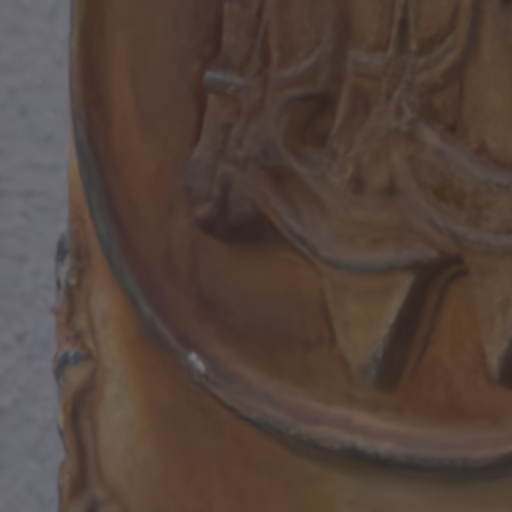}
			\caption*{VDN \cite{VDN} \\ 37.34/0.9619}
		\end{subfigure}			
		\begin{subfigure}[t]{0.24\linewidth}
			\centering
			\includegraphics[width=1\columnwidth]{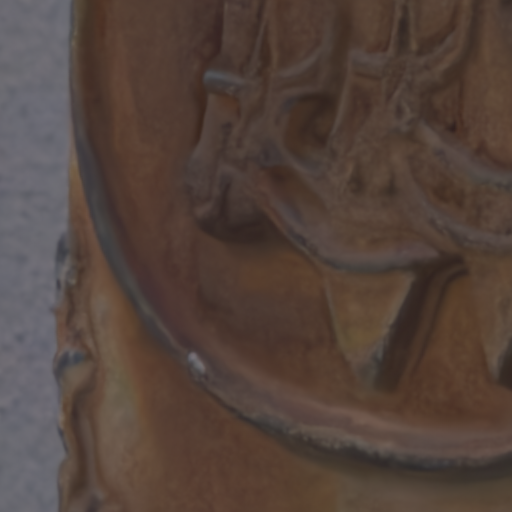}
			\caption*{DUBD-R+ \\ 37.61/0.9637}
		\end{subfigure}	
	\end{center}
	\caption{Visualization of results on a real-noise image from DND \cite{DND}. The PSNR/SSIM results are written below.}
	\label{fig:real}
\end{figure}

\section{Conclusion}
In this paper, we have proposed a universal blind denoiser, which can reduce noise from various environments, including the ones encountered in a real situation. Based on the idea of divide-and-conquer, we split the original denoising MAP problem into two inference sub-problems, and we designed new CNN architectures as sub-problem solvers. Concretely, we introduced an auxiliary random variable to divide and approximate the original problem. Moreover, on the choice of the auxiliary random variable, we explicitly reflected our prior knowledge to augment implicit priors learned from a large-scale dataset. With the experiments, we have shown that our method is an efficient one in terms of performance vs. complexity, and shows promising results in various situations of noise corruption, such as spectrally and spatially varying noises without variance information. As a result, it also works robustly to the real-world noise from cameras. The codes are publicly available at {\fontfamily{pcr}\selectfont \textcolor{mypink}{\url{https://www.github.com/JWSoh/DUBD}}}.

\section*{Acknowledgment}
This research was supported in part by Samsung Electronics Co., Ltd..






\end{document}